\documentclass[10pt,journal,compsoc,final]{IEEEtran}

\usepackage[switch]{lineno} 
\usepackage{dsfont}
\usepackage{bm}
\usepackage{booktabs}
\usepackage{multirow}
\usepackage{arydshln}
\usepackage{xcolor}
\usepackage{array}
\usepackage{float}
\usepackage{eso-pic}

\newcolumntype{Z}{>{\setbox0=\hbox\bgroup}c<{\egroup}@{}}

\ifCLASSOPTIONcompsoc
  \usepackage[nocompress]{cite}
\else
  \usepackage{cite}
\fi
\ifCLASSINFOpdf
   \usepackage[pdftex]{graphicx}
\else
   \usepackage[dvips]{graphicx}
\fi
\usepackage{amsmath}
\ifCLASSOPTIONcompsoc
 \usepackage[caption=false,font=footnotesize,labelfont=sf,textfont=sf]{subfig}
\else
 \usepackage[caption=false,font=footnotesize]{subfig}
\fi
\usepackage{url}
\begin{document}

\title{SegBlocks: Block-Based Dynamic Resolution Networks for Real-Time Segmentation} 

\AddToShipoutPicture*{\scriptsize \sffamily\raisebox{0.7cm}{\hspace{2.4cm}
\begin{minipage}{17cm}
DOI:  10.1109/TPAMI.2022.3162528 \quad \url{https://ieeexplore.ieee.org/document/9744000} \quad © 2022 IEEE. Personal use of this material is permitted. Permission from IEEE must be
obtained for all other uses, in any current or future media, including
reprinting/republishing this material for advertising or promotional purposes, creating new
collective works, for resale or redistribution to servers or lists, or reuse of any copyrighted
component of this work in other works.
\end{minipage}
 }}


% \author{Michael~Shell,~\IEEEmembership{Member,~IEEE,}
%         John~Doe,~\IEEEmembership{Fellow,~OSA,}
%         and~Jane~Doe,~\IEEEmembership{Life~Fellow,~IEEE}% <-this % stops a space
        
\author{Thomas Verelst and~Tinne Tuytelaars%
\IEEEcompsocitemizethanks{\IEEEcompsocthanksitem T. Verelst and T. Tuytelaars are with the Center for Processing Speech and Images, Department Electrical Engineering, KU Leuven.
E-mail: \{thomas.verelst, tinne.tuytelaars\}@esat.kuleuven.be}
}%


\markboth{
%IEEE Transactions on Pattern Analysis and Machine Intelligence
}%
{Verelst \MakeLowercase{\textit{et al.}}: SegBlocks: Block-Based Adaptive Resolution Networks for Real-Time Segmentation}



\IEEEtitleabstractindextext{%
\begin{abstract}
SegBlocks reduces the computational cost of existing neural networks, by dynamically adjusting the processing resolution of image regions based on their complexity. Our method splits an image into blocks and downsamples blocks of low complexity, reducing the number of operations and memory consumption. 
A lightweight policy network, selecting the complex regions, is trained using reinforcement learning. 
In addition, we introduce several modules implemented in CUDA to process images in blocks. Most important, our novel BlockPad module prevents the feature discontinuities at block borders of which existing methods suffer, while keeping memory consumption under control.
Our experiments on Cityscapes, Camvid and Mapillary Vistas datasets for semantic segmentation show that dynamically processing images offers a better accuracy versus complexity trade-off compared to static baselines of similar complexity. For instance, our method reduces the number of floating-point operations of SwiftNet-RN18 by 60\% and increases the inference speed by 50\%, with only 0.3\% decrease in mIoU accuracy on Cityscapes. 
\end{abstract}

}

\maketitle

\IEEEdisplaynontitleabstractindextext

% \fi
\IEEEpeerreviewmaketitle


\IEEEraisesectionheading{\section{Introduction}\label{sec:introduction}}

\IEEEPARstart{C}{omputational} demands of computer vision are constantly increasing, with new deep learning architectures growing in size and new datasets containing images of ever higher resolution. For instance, the Cityscapes dataset for semantic segmentation includes images of 2048 by 1024 pixels~\cite{cordts_cityscapes_2016}, and high-resolution images are also used in medical~\cite{codella_skin_2018} and remote sensing applications~\cite{demir_deepglobe_2018, maggiori_can_2017}. The sheer number of pixels results in a large number of computations due to the convolutions. In addition, many convolutional layers are required to achieve a large receptive field~\cite{chen_deeplab_2018}.

Deep learning applications using these images are often deployed on low-power devices such as phones, surveillance cameras, or car platforms~\cite{canziani_analysis_2017,sze_efficient_2017}. The interest in embedded computer vision has lead to efficient neural network architectures~\cite{sandler_mobilenetv2_2018, ma2018shufflenet, ferrari_icnet_2018, romera_erfnet_2018} and model compression methods~\cite{lecun_optimal_1990, hinton_distilling_2015,wu_quantized_2016}.

\begin{figure}[tb]
\centering
\includegraphics[width=0.95\linewidth]{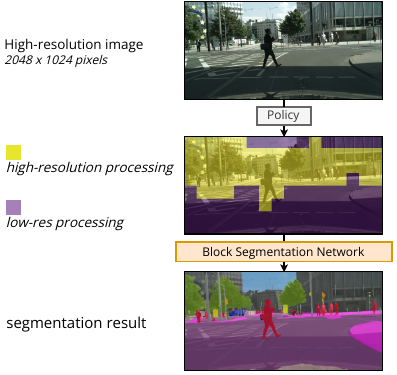}
\caption{SegBlocks adjusts the processing resolution of image regions based on their complexity. 
A lightweight policy network decides which blocks should be processed in high resolution mode. The number of operations is reduced without significant loss of accuracy.
\label{fig:overview}}
\end{figure}

These approaches result in static neural networks, which apply identical operations to every pixel.
%in an image
Yet not every image region is equally complex. Therefore, static networks may not be allocating operations in the most optimal way. 
% For instance, sky and vegetation regions are easy to segment in comparison to traffic signs or cars. 
Various methods have been proposed to dynamically adapt network architectures based on image complexity.
Most are designed for classification tasks, for instance by completely skipping network layers or channels~\cite{ferrari_skipnet_2018, veit2018convolutional, wu_blockdrop_2018, bejnordi_batch-shaping_2020, verelst_dynamic_2020} and only few methods target dense pixel-wise classification tasks~\cite{huang_uncertainty_2019,wu_patch_2020, wu_real-time_2017}. Dynamic methods typically focus on the theoretical reduction in computational complexity, due to the implementation challenges of sparse operations~\cite{li_not_2017, bejnordi_batch-shaping_2020,figurnov_spatially_2017,verelst_dynamic_2020}. 
% For example, methods relying on sparse convolutions only demonstrate speedup on CPU~\cite{li_not_2017, bejnordi_batch-shaping_2020}, on specific operations~\cite{verelst_dynamic_2020}, or demonstrate no inference speed improvements~\cite{figurnov_spatially_2017}. 
%In this work, we achieve inference speedup on GPU using block-based processing. %~\cite{ren_sbnet_2018}.
In this work, we speed up inference using block-based processing, of which efficient implementations have already been demonstrated on GPU and other platforms~\cite{vooturi_dynamic_2019 ,ren_sbnet_2018, sun2020computation}.

We propose a method to dynamically process low-complexity regions at a lower resolution, as illustrated in Figure~\ref{fig:overview}. Our method splits an image into blocks, and then downsamples non-important blocks. Reducing the processing resolution not only reduces the computational cost, but also decreases the memory consumption. % of these regions. 
The policy network, selecting the regions to be processed in high resolution, is trained using reinforcement learning.

\begin{figure}[!tb]
\centering
\subfloat[Convolution on full image]{\includegraphics[width=0.4\linewidth, trim=0 23 37 0, clip]{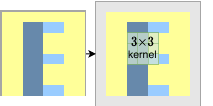}
\label{fig:conv_std}}
\hfil
\subfloat[Convolution on blocks]{\includegraphics[width=0.4\linewidth, trim=0 24 40 0, clip]{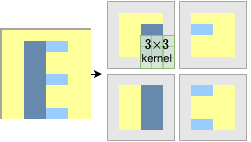}
\label{fig:zeropad}}
\caption{Illustration of the zero-padding problem when processing blocks: convolutions pad individual blocks with zeros (grey), 
stopping the propagation of features between blocks and therefore resulting in a loss of global context.}
\label{fig:patch_problem}
\end{figure}

Efficiently processing images in blocks without losing accuracy is not trivial. One could treat each block as a separate image and then combine the individual block outputs. However, features cannot propagate between individual blocks (Figure~\ref{fig:zeropad}), leading to a loss in global context and significant decrease in accuracy. Existing works partially address this by including global features extracted by a separate network branch, requiring custom architectures~\cite{wu_patch_2020, wu_real-time_2017}.

\begin{figure}[!tb]
\centering
\includegraphics[width=1\linewidth]{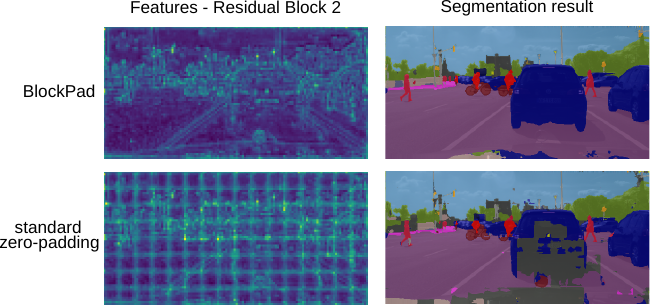}
\caption{Comparison of feature maps when processing an image in blocks with either standard zero-padding or our custom BlockPad module. Features are visualized by summing activation magnitudes over the channel dimension and tiling the outputs of individual blocks into a single image. %, for outputs of three residual blocks in the network.
Standard zero-padding introduces noticeable artifacts at block borders. In addition, the output shows inconsistencies, e.g. in the car, due to the lack of global context for individual blocks. In contrast, our BlockPad module enables feature propagation as if the network was never evaluated in blocks. }
\label{fig:features}
\end{figure}

Our method addresses this problem by replacing zero-padding with our BlockPad operation. This custom CUDA module copies features from neighboring blocks into the padding border and therefore preserves feature propagation between blocks, as if the network was never split into blocks. A comparison between zero-padding and the BlockPad module is given in Figure~\ref{fig:features}, showing that our module avoids artifacts at block borders.
% Whereas standard zero-padding introduces artifacts in the feature maps, BlockPad 
% With BlockPad, we can efficiently process existing network architectures in blocks without adverse effects.

The contributions of this work are as follows:
\begin{itemize}
    \item We introduce the concept of dynamic block-based convolutional neural networks, where blocks are downsampled based on their complexity, to reduce their computational cost. In addition, we provide CUDA modules for PyTorch to efficiently implement block-based methods \footnote{The code is available at \newline
\mbox{\footnotesize{\url{https://github.com/thomasverelst/segblocks-segmentation-pytorch}}} }.
    
    \item We train the policy network with reinforcement learning, in order to select complex regions for high-resolution processing.
    
    \item We demonstrate our method using a state of the art semantic segmentation network, and show that our method reduces the number of floating-point operations~(FLOPS) and increases the inference speed~(FPS) with only a slight decrease in mIoU accuracy. Our method achieves better accuracy than static baseline networks of similar complexity.
\end{itemize}

This work is an extension of a 4-page short paper~\cite{verelst2020segblocks}, where block-based processing was introduced in combination with a simple heuristic-based downsampling policy that selects regions where the loss of visual detail is most significant. 
Here, we provide more details on the block-based processing framework and we introduce a superior reinforcement learning downsampling policy that is trained specifically for the task at hand. In addition, we provide results for Cityscapes, CamVid and Mapillary Vistas datasets with more comprehensive ablation studies.

\section{Related work}

\subsection{Semantic segmentation}

Traditional works in semantic segmentation
focus on improving segmentation accuracy, without taking into account the network complexity~\cite{zhao_pyramid_2017, chen_deeplab_2018, badrinarayanan_segnet_2017}. In contrast to other tasks such as classification or object detection, segmentation requires pixel-wise labels of the same resolution as the input. Preserving spatial information in the network requires high-resolution latent representations with a high computational and memory cost. At the same time, a large receptive field is needed to incorporate global context. In order to keep the network size under control, semantic segmentation networks use encoder-decoder architectures with skip connections~\cite{badrinarayanan_segnet_2017}, dilated convolutions~\cite{yu2015multi} and spatial pyramid pooling~\cite{chen_deeplab_2018}.

Applications such as driverless cars have raised interest in real-time inference on embedded devices. Several high-resolution datasets for these applications have emerged.
The Cityscapes dataset~\cite{cordts_cityscapes_2016} provides images of $2048{\times}1024$ pixels, with highly detailed annotations for 30 semantic classes. A more extensive dataset is Mapillary Vistas~\cite{neuhold_mapillary_2017} with 25000 high-resolution images and 152 object categories. As real-time inference is crucial for these applications and datasets, smaller and more efficient segmentation networks have been developed for this task specifically. 

ICNet~\cite{ferrari_icnet_2018} proposes a custom encoder architecture processing an image pyramid, with multi-resolution feature maps fused before the decoder. ERFNet~\cite{romera_erfnet_2018} factorizes convolution kernels into $1{\times}3$ and $3{\times}1$ kernels to reduce the computational cost. Bilateral Segmentation Network~(BiSeNet)~\cite{ferrari_bisenet_2018} presents a network with two branches: a Spatial Path to encode high-resolution spatial information and a Context Path to achieve a high receptive field. A similar dual-branch architecture is proposed by Guided Upsampling Network (GUN)~\cite{mazzini2018guided}, where both branches share weights.
A multi-branch network at different scales is demonstrated by ShelfNet~\cite{zhuang2019shelfnet}.
DABNet~\cite{li2019dabnet} proposes a Depthwise Asymmetric Bottleneck module, which combines dilated convolutions with non-dilated ones, to capture local and more contextual information. ESPNetv2~\cite{mehta2019espnetv2} introduces a building block with dilated depthwise convolutions for large receptive fields.
Other methods use attention modules~\cite{huang2019ccnet, wu_dynamic_2020, zhu_asymmetric_2019, hu_real-time_2020} to incorporate global context, for instance by weighting spatial features~\cite{wang_non-local_2018}.

Many segmentation networks use classification backbones. SwiftNet~\cite{orsic_defense_2019} demonstrates that state of the art results can be achieved with a straightforward architecture: a \mbox{ResNet-18} encoder~\cite{he_deep_2016}, pretrained on ImageNet~\cite{deng_imagenet_2009}, is combined with a spatial pyramid pooling (SPP) module~\cite{he_spatial_2015} and basic decoder. 
Lightweight backbones such as MobileNetV2~\cite{sandler_mobilenetv2_2018}, ShuffleNetV2~\cite{ma2018shufflenet} and EfficientNet~\cite{tan2019efficientnet} use depthwise convolutions to reduce the computational cost, leading to models such as SwiftNetMN-V2~\cite{orsic_defense_2019}.
Model compression techniques can further reduce the computational cost, by applying pruning~\cite{lecun_optimal_1990}, knowledge distillation~\cite{hinton_distilling_2015}, quantization~\cite{wu_quantized_2016}, or factorization~\cite{sainath_low-rank_2013,jaderberg_speeding_2014}. 

\subsection{Dynamic segmentation}
These efficient architectures and compression techniques achieve impressive performance, but do not exploit spatial properties of the image as they process each pixel with the same operation. Some recent methods address this by processing the image dynamically in blocks or patches. It is worth noting that dynamic methods are often complementary to static acceleration methods, and both can be combined to further reduce the computational cost.

The method proposed by Huang~\emph{et al.}~\cite{huang_uncertainty_2019} combines a fast segmentation model with a large model. First, the image is processed by a fast network, such as ICNet~\cite{ferrari_icnet_2018}. Then, a selection criterion, based on the softmax {certainty} of the output,
uses the rough predictions to decide which image regions should be re-evaluated by the large model (e.g. PSPNet~\cite{zhao_pyramid_2017}). % for more accurate segmentations. 
Such a two-stage approach has two drawbacks. First, the large network processes the image in blocks and the discontinuities at block borders stop feature propagation. Therefore, the method uses large patches of at least $256{\times}256$ pixels to include some global context. Secondly, the feature maps and predictions of the small network are not re-used by the large network, making the method less efficient and only applicable when the accuracy and cost difference between the small and large network is substantial. 

Wu~\emph{et al.}~\cite{wu_real-time_2017} propose a custom architecture where a high-resolution branch improves rough predictions of a low-resolution branch. Again, this method struggles with the feature propagation problem at patch borders and therefore uses large blocks of $256{\times}256$ pixels. In addition, they incorporate global features from the low-resolution network into the high-resolution branch to provide more global context. The Patch Proposal Network~\cite{wu_patch_2020} method of Wu~\emph{et al.} has a similar architecture, with a low-resolution global branch and high-resolution refinement branch. The local refinement branch only operates on selected patches, based on a trained selection criterion where regions with below-average accuracy are refined.  Features of both branches are fused before generating final predictions. 

In contrast to these methods, our dynamic method does not require modifications to existing network architectures. We address the discontinuities at block borders without requiring additional global features in the patch-based processing. Furthermore, by being applicable on existing architectures, our method benefits from further advancements in network architectures. 

Another approach to reduce spatial redundancy is to warp the input image to enlarge important regions~\cite{recasens2018learning}. The work of Marin~\emph{et al.}~\cite{marin_efficient_2019} uses non-uniform image sampling to focus on complex regions. However, as the internal representation of the network is still a regular-spaced pixel grid, the flexibility of non-uniform downsampling is limited, and the network typically focuses on a single region. 

\subsection{Conditional execution}

Adapting the network architecture based on the input image is also known as \textit{conditional execution}~\cite{bengio_conditional_2016,bengio_estimating_2013,bengio_deep_2013} or dynamic neural networks. For instance, some methods reduce the processing cost of simple images by skipping complete residual blocks~\cite{ferrari_skipnet_2018,wu_blockdrop_2018,veit2018convolutional} or by dynamically pruning feature channels~\cite{bejnordi_batch-shaping_2020}. However, these methods are intended for classification tasks, with a large variety in features between images of different classes. In segmentation, many images contain the same objects and features in different spatial arrangements. 

Some dynamic methods rely on spatial properties to reduce the computational cost: they skip computations on low-complexity regions. Spatially Adaptive Computation Time~\cite{figurnov_spatially_2017} and cascade-based methods~\cite{li_not_2017} halt the computation of features when features are `good enough'. 
DynConv~\cite{verelst_dynamic_2020} demonstrates inference speed improvements on human pose estimation tasks, where large regions of the image can be completely ignored by the network.  The methods of Xie~\emph{et al.}~\cite{xie_spatially_2020} and Figurnov~\emph{et al.}~\cite{figurnov2016perforatedcnns} dynamically interpolate features in low-complexity regions. Requiring sparse convolutions, these methods either demonstrate no inference speed improvements~\cite{figurnov_spatially_2017}, only on depthwise convolutions~\cite{verelst_dynamic_2020}, or only on CPU~\cite{xie_spatially_2020, figurnov_spatially_2017}, or are intended for specialized hardware~\cite{hua2019channel}. Block-based approaches are considered to be more feasible to implement efficiently on most platforms~\cite{vooturi_dynamic_2019 ,ren_sbnet_2018, sun2020computation}. In addition, methods completely skipping non-complex regions are less suitable for pixel-wise labeling tasks such as segmentation. Our method processes every image region, but reduces the cost of non-complex regions.

An important aspect of conditional execution is the policy which determines the layers, channels or regions to execute. Often, this policy cannot be trained by standard backpropagation due to its discrete nature.  Recently, the Gumbel-Softmax~\cite{jang_categorical_2017,maddison_concrete_2017} trick gained popularity in dynamic methods~\cite{veit2018convolutional, verelst_dynamic_2020, bejnordi_batch-shaping_2020}. In our case, it would require predictions for each region in both high and low-resolution, increasing the computational cost at training time. Instead, we show that the policy can be trained efficiently using reinforcement learning~\cite{sutton_reinforcement_1998}.

\begin{figure*}[!tb]
\centering
\includegraphics[width=\textwidth,trim= 0 0 50 0]{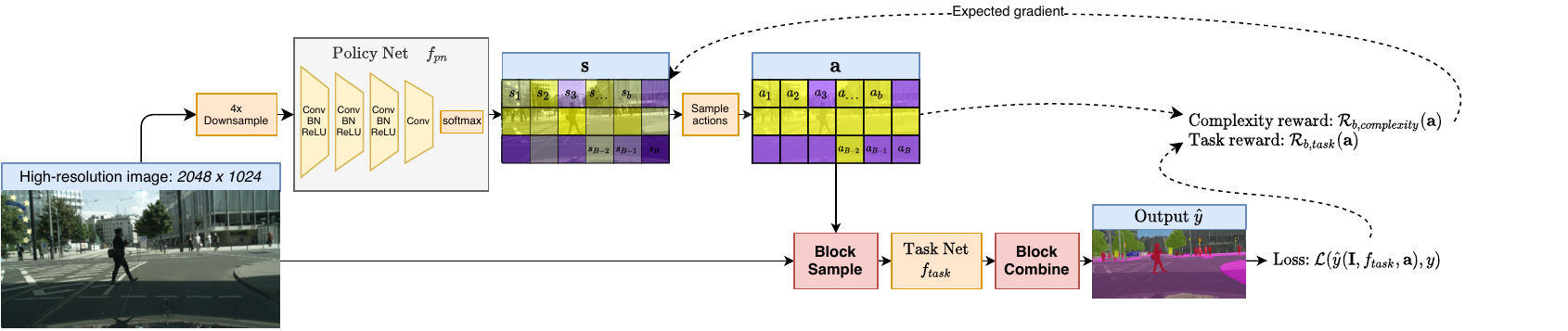}
\caption{Training the policy network with reinforcement learning: The lightweight policy net predicts soft decisions $\bm s$, which are  sampled to actions $\bm a$. The policy net's gradients are estimated using REINFORCE~\cite{williams1992simple} with a separate reward per block, based on the task loss in a block's region and the number of blocks processed at high resolution.}
\label{fig:arch_rl}
\end{figure*}

\section{Method}

SegBlocks splits an image into blocks and adjusts the processing resolution of each block based on its complexity. A lightweight policy network processes a low-resolution version of the image and outputs resolution decisions for each image region. For implementation simplicity, we restrict the resolution to two options, with either high- or low-resolution processing. Note that high- and low-resolution regions are processed by the same convolutional filters, which consequently perceive objects at different scales. This does not affect the performance negatively though, as segmentation images typically have large variations in object size, and convolutional neural networks are trained to be robust against scale variations.
Since resolution decisions are discrete, the policy network is trained with reinforcement learning, using a reward function to determine the expected gradient. % of the policy network.

First, we discuss how the reinforcement learning is incorporated using a reward taking into account both computational complexity and task accuracy. The policy is trained jointly with the main segmentation network. In the second part, we elaborate on custom modules making block-based adaptive processing of images possible.

\subsection{Downsampling policy}

\subsubsection{Policy network}
The policy network is a convolutional neural network that outputs a discrete decision per block. The network should be small compared to the main segmentation network.
We use a simple architecture with 4 convolutional layers of 64 channels, batch normalization and ReLU non-linearities followed by a softmax layer over the channel dimension. A four times downsampled version of the image is given as input.
The policy network $f_{pn}$, parametrized by $\theta$, outputs probabilities $s_b$ per block, indicating the likelihood that the block should be processed in high resolution based on the input image $\bm I$:
\begin{equation}
\bm s = f_{pn} ( \bm I ; \theta ), 
\end{equation}
\begin{equation}
\text{with } \bm s = [s_1, \dots, s_b, \dotsc, s_B] \in [0,1]^B.
\end{equation}
The soft probabilities are sampled to actions $\bm a$,
where \mbox{$a_b=1$} results in high-resolution processing of block $b$:
\begin{equation}
s_b = P(a_b = 1).
\end{equation}
Standard backpropagation cannot be applied due to the sampling of discrete actions. Finding optimal decisions $\bm a$ for context $\bm I$, with a constrained budget, is also referred to in literature as contextual bandits~\cite{sutton_reinforcement_1998}. This can be seen as a reinforcement learning scenario with a single time step: the policy network predicts all actions $a_b$ at once and directly obtains the reward.
The probability of actions $\bm a$, for image $\bm I$ and parameters $\theta$ is given by
\begin{equation}
\pi_{\theta}( \bm a \mid \bm I) = \prod_{b=1}^{B} p_{\theta}( a_b \mid \bm I).
\end{equation}
The objective is to find policy parameters $\theta$, so that the predicted actions $\bm a$ for image $\bm I$ maximize the policy reward.

\subsubsection{Policy reward}
The goal of the policy network is to select the most important blocks for high-resolution processing. To avoid early convergence to a sub-optimal state where all blocks are processed at high resolution, the reward takes into account both the number of blocks processed at high resolution and the task accuracy. Figure~\ref{fig:arch_rl} illustrates how both rewards are integrated in the training process.

We assume that a resolution decision $a_b$ only affects the segmentation output of block $b$ and does not influence adjacent blocks. The total reward for an image can then be expressed as the sum of individual block rewards. A block reward consists of one term optimizing accuracy and second term keeping the number of operations under control, weighted by hyperparameter $\gamma$ (set to 10 in our experiments). The reward per image is then written as
\begin{equation}
\begin{split}
    \mathcal{R}(\bm a) &= \sum_b^B{\mathcal{R}_b(\bm a)} \\
    &= \sum_b^B{ \bigl(   \mathcal{R}_{b,task}(\bm a) + \gamma \mathcal{R}_{b,complexity}(\bm a)     \bigr)} .
\end{split}
\end{equation}
The percentage of blocks processed in high-resolution is given by
\begin{equation}
  \sigma(\bm a) = \frac{1}{B}\sum^{B}_{b=1}{a_b} .
\end{equation}
Then, the following reward minimizes 
the difference between $\sigma$ and the desired percentage $\tau$:
\begin{equation}
    \mathcal{R}_{b,complexity}(\bm a) = 
    \begin{cases}
     -( \sigma(\bm a) - \tau) &\text{ if } a_b = 1, \\
     \sigma(\bm a) - \tau &\text{ if } a_b = 0 .
    \end{cases}
    \label{eq:reward_complexity}
\end{equation}
This reward is positive for blocks processed at high resolution when the actual percentage $\sigma$ is smaller than the desired percentage $\tau$, resulting in an incentive for the policy network to process more blocks at high resolution. Using a target $\tau$ instead of simply minimizing $\sigma$ results in more stable training and lower sensitivity to hyperparameter $\gamma$.

The task reward encourages the policy network to select regions with a high segmentation loss for high resolution processing. In general, those regions correspond to the most complex ones in the image. The task criterion, e.g. pixel-wise cross entropy, is denoted by $ \mathcal{L}$. The task loss per image is then given by 
\begin{equation}
{L}_{task} = \mathcal{L}(\mathbf{ \widehat{y}}, \mathbf{y}) 
\end{equation}
where $\mathbf{\widehat{y}}$ and  $\mathbf{y}$ denote the predictions and ground truth labels respectively. Our reward is based on the task loss per block. Values $\mathbf{\widehat{y}}_b$ and  $\mathbf{y}_b$ are obtained by only considering values in the block region. The task reward of block $b$ can then be defined as
\begin{equation}
    \mathcal{R}_{b,task}(\bm a) =
    \begin{cases}
     \mathcal{L}( \mathbf{\widehat{y}}_b, \mathbf{y}_b) - {L}_{task}     & \text{ if } a_b = 1, \\
    - (\mathcal{L}( \mathbf{\widehat{y}}_b, \mathbf{y}_b)  - {L}_{task})    &\text{ if } a_b = 0.
    \end{cases}
\end{equation}
Subtracting ${L}_{task}$ is not strictly necessary but reduces the variance of the rewards for more stable training.

\subsubsection{Expected gradient}

The policy network should predict actions that maximize the expected reward:
\begin{equation}
\max \mathcal{J}(\theta) =\max \mathds{E}_{\bm a \sim \pi_{\theta}} [\mathcal{R}(\bm a)].
\end{equation}
The policy network's parameters ${\theta}$ can then be updated using gradient ascent with learning rate $\alpha$:
\begin{equation}
{\theta} \leftarrow {\theta}  + \alpha \nabla_{{\theta}} [\mathcal{J}({\theta})]
\end{equation}
The gradients of $J$ can be derived similarly to the policy gradient method REINFORCE~\cite{williams1992simple}:
\begin{equation}
\label{eq:reinforce}
\begin{split}
\nabla_{\theta}\mathcal{J}(\theta) &= \nabla_{\theta}\mathds{E}_{\bm a} [\mathcal{R}(\bm a)]\\
&= \nabla_{\theta} \sum_{\bm{a}} { \pi_{\theta}( \bm a \mid \bm I) \mathcal{R}(\bm a) }\\
&= \sum_{\bm{a}} {  \nabla_{\theta} \pi_{\theta}( \bm a \mid \bm I) \mathcal{R}(\bm a) }\\
&= \sum_{\bm{a}} { \pi_{\theta}( \bm a \mid \bm I) \nabla_{\theta}[{\log\pi_{\theta}( \bm a \mid \bm I)}\mathcal{R}(\bm a) }]\\
&= \sum_{\bm{a}} { \pi_{\theta}( \bm a \mid \bm I) \nabla_{\theta} \Bigl[{\log \prod_{b=1}^{B} p_{\theta}( a_b \mid \bm I) } \Bigr] \mathcal{R}(\bm a) }\\
&= \sum_{\bm{a}} { \pi_{\theta}( \bm a \mid \bm I)  \mathcal{R}(\bm a) \sum_{b=1}^{B} \bigl( \nabla_{\theta}{  \log  p_{\theta}( a_b \mid \bm I) }\bigr) }\\
&= \mathds{E}_{\bm a}\Bigl[ { \mathcal{R}(\bm a)  \sum_{b=1}^{B}\bigl( \nabla_{\theta}{  \log  p_{\theta}( a_b \mid \bm I) }\bigr) }\Bigr]\\
&= \mathds{E}_{\bm a}\Bigl[ { \sum_{b=1}^{B} \bigl( \mathcal{R}_b(\bm a) \nabla_{\theta}{  \log  p_{\theta}( a_b \mid \bm I) }} \bigr)\Bigr] .
\end{split}
\end{equation}
In practice, the expectation in Equation~\ref{eq:reinforce} is approximated with Monte-Carlo sampling using samples in the mini-batch. The result can be interpreted as applying REINFORCE on each block individually with reward $\mathcal{R}_b(\bm a)$, giving unbiased but high-variance gradient estimates. In practice,  we found the policy network stable to train for a large range of hyperparameters.

The segmentation network and policy network are jointly trained. The hybrid loss to be minimized is then given by
\begin{equation}
{L}_{hybrid} = {L}_{task} + \frac{\beta}{N}  \sum_{n=1}^{N} { \sum_{b=1}^{B} \bigl(-\mathcal{R}_b(\bm a) {  \log  p_{\theta}( a_b \mid \bm I) }} \bigr)
\end{equation}
with $N$ the batch size and $\beta$ a hyperparameter weighting the loss terms (set to 4 in our experiments). 

\subsection{Block modules}

%%%%% PLACED FORWARD FOR LATEX FIGURE PLACEMENT
\begin{figure*}[!tb]
\centering
\includegraphics[width=\textwidth]{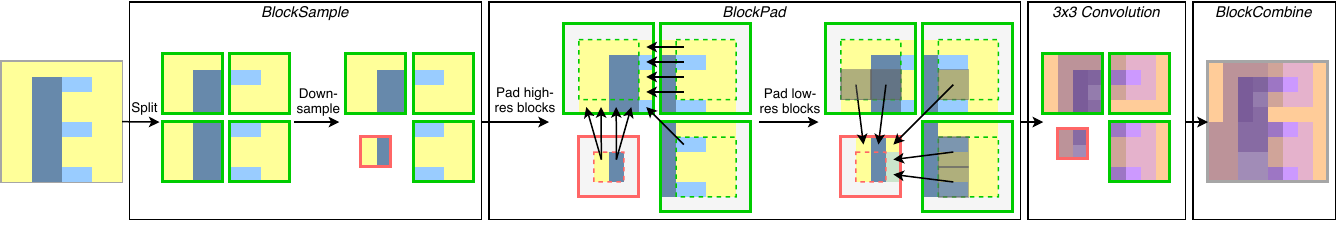}
\caption{Illustration of the BlockSample, BlockPad and BlockCombine modules. BlockSample splits the image and downsamples low-complexity blocks. BlockPad replaces zero-padding and enables feature propagation between blocks.
% , which is typically used to preserve the spatial dimensions after a convolution. 
The module copies features from neighboring blocks into the padding. When padding low-resolution blocks adjacent to high-resolution blocks, multiple pixel values of the high-resolution blocks are averaged to better preserve the spatial coherence of features.
\label{fig:blockpad}}
\end{figure*}

We introduce three new modules for block-based processing with convolutional neural networks: BlockSample, BlockPad, and BlockCombine. Figure~\ref{fig:blockpad} gives an overview of a simple block-processing pipeline.

\subsubsection{BlockSample}
The BlockSample module splits the image into blocks and downsamples low-complexity blocks consecutively. Average pooling is used for efficient downsampling. We use a downsampling factor of 2, reducing the computational cost of low-complexity regions by a factor 4.
The CUDA implementation fuses the splitting and downsampling steps into a single operation. Images are split into blocks of a predefined block size, for instance $128{\times}128$ pixels. Downsampling operations, such as max pooling, further reduce the spatial dimensions of blocks throughout the network.
% The output is a stack of high-resolution regions and a stack of low-resolution regions, which together form a BlockRepresentation.

% low-complexity regions are downsampled by a factor 2 to reduce their computational cost. We use blocks of $128{\times}128$ pixels. As the network pools down feature maps, the block size is reduced accordingly up to just $4{\times}4$ pixels in the deepest layers, offering fine granularity.

\subsubsection{BlockPad}
\label{sec:blockpad}

The BlockPad module replaces zero-padding and eliminates discontinuities at block borders by copying features from neighboring blocks into the padding.  Evaluating the network with only high-resolution blocks is equivalent to normal processing without blocks.

\begin{figure}[!tb]
\centering
\subfloat[Strided subsampling]{\includegraphics[width=0.35\linewidth]{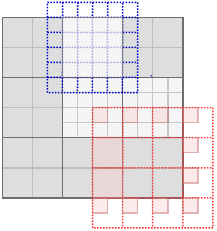}%
\label{fig:sampling_stride}}
\hfil
\subfloat[Average sampling]{\includegraphics[width=0.35\linewidth]{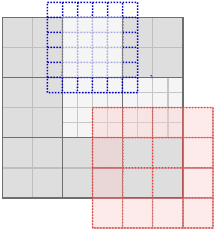}%
\label{fig:sampling_avg}}
\caption{BlockPad aims to respect the spatial relationship between high- and low-resolution blocks by sampling using the illustrated patterns. Low-resolution blocks ($2{\times}2$ pixels size) are colored dark grey, while the $4{\times}4$ high-resolution blocks are colored light grey. The dotted blue and red grid shows the sampling pattern when padding high-resolution and low-resolution blocks respectively.}
\label{fig:sampling}
\end{figure}

When two high-resolution or two low-resolution blocks are adjacent, pixels can be copied directly while preserving their spatial relationship. When copying features from low-resolution blocks into the padding of high-resolution blocks, features are nearest-neighbor upsampled in order to preserve the spatial relationship, as illustrated in Figure~\ref{fig:blockpad}. When copying features from high-resolution blocks into the low-resolution block's padding, we considered two possibilities: strided subsampling and average sampling. Figure~\ref{fig:sampling_stride} illustrates that subsampling copies features in a strided pattern, whilst average sampling (Fig~\ref{fig:sampling_avg}) combines multiple pixels. Our experiments show that average sampling achieves better results with a small increase in overhead.

\subsubsection{BlockCombine}
The BlockCombine module upsamples low-resolution blocks to the same resolution as the high-resolution ones, and then combines all blocks into a single tensor. Our CUDA implementation merges the nearest neighbour upsampling and block combining steps to reduce the overhead of this operation.

\section{Experiments}
We integrate our dynamic processing method in SwiftNet~\cite{orsic_defense_2019}, a state of the art network for real-time semantic segmentation of road scenes. We test semantic segmentation on the Cityscapes~\cite{cordts_cityscapes_2016}, Camvid~\cite{brostow2009semantic_camvid}, and Mapillary Vistas~\cite{neuhold_mapillary_2017} datasets. In addition, we provide an ablation study, analyzing the overhead of block-based processing. Our method is implemented in PyTorch 1.5 and CUDA 10.2, without TensorRT optimizations. We report two model complexity metrics: the number of computations and the frames per second.

{The number of computations} is reported in billions of \emph{multiply-accumulates (GMACs)} per image, being half the number of floating-point operations (FLOPS). The FLOPS metric is often used to compare model complexity in a platform-agnostic manner, since the efficiency of the implementation has no impact. 
For our dynamic method, where the number of computations varies per image, we report the average GMAC count, over the validation or test set.

\emph{Frames per second (FPS)} is a more practical metric to measure inference speed, but depends largely on the implementation of building blocks. 
% For instance, 
% For instance, depthwise convolutions offer only a slight inference speed improvement over standard convolutions on GPUs, even though the number of computations is much lower. Still, some backbones use depthwise convolutions due to their efficiency on mobile platforms~\cite{sandler_mobilenetv2_2018}. 
Therefore, both FPS and GMACs metrics should be taken into account for fair comparison. 
We measure the inference speed on both a high-end Nvidia GTX 1080 Ti 11GB GPU and low-end Nvidia GTX 1050 Ti 4 GB GPU, paired with an Intel i7 processor. The model is warmed up on the first half of the image set, and the speed is measured on the second half. To compare the inference speed to those reported by others, we normalize the FPS numbers (\textit{norm FPS}) of different GPUs based on their relative performance, using the same scaling factors as Orsic \emph{et al.}~\cite{orsic_defense_2019}.

\subsection{Cityscapes semantic segmentation}

\begin{figure}[tb]
\centering
\includegraphics[width=1\linewidth, trim= 0 0 0 0, clip]{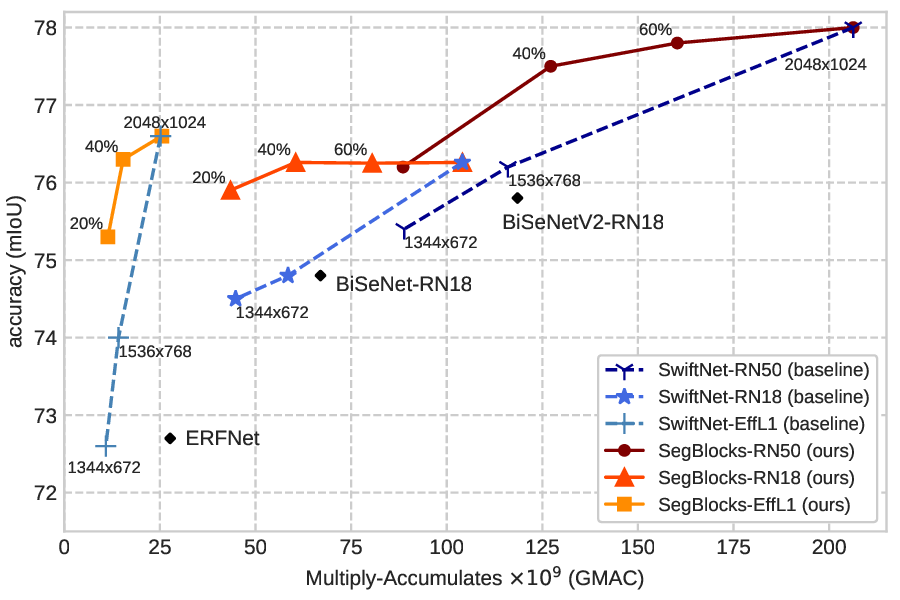}
\caption{Cityscapes validation results and comparison with static baselines of similar complexity. For a given computational complexity (GMACs), our adaptive resolution method SegBlocks consistently outperforms static baseline networks (SwiftNet) of similar complexity. Each backbone (RN50, RN18, Eff-L1) is trained and evaluated with $\tau \in {0.2,0.4,0.6}$ for dynamic models and resolutions $2048{\times}1024$, $1536{\times}768$, $1344{\times}672$ for static baselines.  }
\label{fig:cityscapes_results}
\end{figure}

\begin{table*}[tb]
\caption{Results on Cityscapes semantic segmentation. Our SegBlocks models are based on the respective SwiftNet baselines, and integrate block-based dynamic resolution processing in those networks \label{tab:cityscapes_results}. {The symbol `-' indicates that the metric was not reported. For our method, only several test set results are reported due to submission limitations on Cityscapes test evaluation.}}
\centering
\begin{tabular}{@{}clcccccZ@{}}
% \setlength{\tabcolsep}{2pt}
% \scriptsize
\toprule
& \textbf{Method}                & \textbf{mIoU \textit{val}} & \textbf{\textit{test}} & \textbf{GMACs}  & \textbf{FPS} & \textbf{norm FPS} & \textbf{memory (max)} \\ \midrule

\multirow{9}{*}{\rotatebox[origin=c]{90}{Our method}} &
SwiftNet-RN50 (baseline, our impl.)               & 78.0 &           & 206.3     &     16.3     @ 1080 Ti,\quad  3.8  @ 1050 Ti         &   16.3          & 2724  MB            \\
& SegBlocks-RN50 ($\tau$=0.4)  & 77.5 &       76.1           &  127.2  &    23.3   @ 1080 Ti,\quad 5.7 @ 1050 Ti       &           23.3       &    2957  MB        \\
& SegBlocks-RN50 ($\tau$=0.2)  & 76.2 &                  &  88.5  &    30.0    @ 1080 Ti,\quad 7.3 @ 1050 Ti       &            30.0        &    2044 MB            \\[0.5ex]
\cdashline{2-8}\noalign{\vskip 0.5ex}
 & SwiftNet-RN18 (baseline, our impl.)               & 76.2 &   74.4        & 104.1     &      38.4   @ 1080 Ti,\quad   9.9 @ 1050 Ti         &      38.4       &   480/717 2256/1321 MB            \\
 & SegBlocks-RN18 ($\tau$=0.4)            &  76.3 &    74.6      & 60.5                 &      48.6     @ 1080 Ti,\quad   13.5 @ 1050 Ti         &    48.6                &  519/1386 1862  MB            \\
 & SegBlocks-RN18 ($\tau$=0.2)            &  75.9 &         & 43.5                  &       57.8    @ 1080 Ti,\quad  16.8 @ 1050 Ti        &      57.8              &   463 / 1384  MB            \\[0.5ex]
\cdashline{2-8}\noalign{\vskip 0.5ex}

 & SwiftNet-EffL1 (baseline, our impl.)               & 76.6 &           & 24.9     &      17.4   @ 1080 Ti,\quad   7.1 @ 1050 Ti         &       17.4     &   1338 MB            \\
 & SegBlocks-EffL1 ($\tau$=0.4)            &  76.3 &  74.1        & 15.6                 &        30.4   @ 1080 Ti,\quad  8.4 @ 1050 Ti         &        30.4          &  1210   MB            \\
 & SegBlocks-EffL1 ($\tau$=0.2)            &  75.3 &              & 11.4                 &      35.6     @ 1080 Ti,\quad  10.6 @ 1050 Ti         &        35.6          &  1210   MB            \\
%  & SegBlocks-SNv2 ($\tau$=0.2)            &  74.5 &         & 20.6                  &      56.0  @ 1080 Ti,\quad  16.1 @ 1050 Ti        &      56.0              &   1192  MB            \\
\midrule

\multirow{5}{*}{\rotatebox[origin=c]{90}{\parbox{1.1cm}{\centering Dynamic\\networks}}} & Patch Proposal Network (AAAI2020)~\cite{wu_patch_2020}                    & 75.2 &         -  &  -           & 24 @ 1080 Ti  & 24               &   1137 MB               \\
 & Huang et al. (MVA2019)~\cite{huang_uncertainty_2019}  & 76.4 &     -      & -           & 1.8 @ 1080 Ti   & 1.8                &        -           \\
 & Wu et al.~\cite{wu_real-time_2017}    & 72.9 &      -     & -           & 15 @ 980 Ti   & 33    & -                  \\ 
 & Learning Downsampling (ICCV2019)~\cite{marin_efficient_2019}    & 65.0 &   -        &  34           & -   &-    & -                  \\ 
 & HyperSeg-M~\cite{nirkin2021hyperseg} & 76.2 & 75.8 & 7.5 & 36.9 @ 1080 Ti & 36.9 & - MB \\
 & Stochastic Sampling~\cite{xie_spatially_2020} & 80.6 & - & 373.2 & no GPU implementation &- & - \\
\midrule

\multirow{8}{*}{\rotatebox[origin=c]{90}{\parbox{2cm}{\centering Efficient\\static networks}}}
 & ShelfNet18-lw (CVPR2019)~\cite{zhuang2019shelfnet}     & - &      74.8     & 95        & 36.9 @ 1080 Ti  & 36.9               &      -           \\
 & SFNet (ResNet-18) (ECCV2020)~\cite{zhuang2019shelfnet}     & - &      78.9     & 247        & 26 @ 1080 Ti  & 26               &      -           \\
 & SwiftNet-RN18 (CVPR2019)~\cite{orsic_defense_2019}     & 75.6 &      75.5     & 104.0        & 39.9 @ 1080 Ti  & 39.9               &      -           \\
  & ESPNetV2 (CVPR2019)~\cite{mehta2019espnetv2}     & 66.4 & 66.2          & 2.7        & -  & -               &      -           \\
 & DABNet (BMVC2019)~\cite{li2019dabnet}  & 70.1 &      -     & -        & 27.7 @ 1080 Ti  & 27.7               &      -           \\
 & BiSeNet-RN18 (ECCV2018)~\cite{ferrari_bisenet_2018}     & 74.8 &    74.7        & 67       & 65.5 @ Titan Xp & 58.5               &        -       \\
 & BiSeNetV2-RN18 (IJCV2021)~\cite{yu2021bisenet}     & 75.8 &    75.3        & 118.5       & 47.3 @ 1080 Ti & 47.3               &        -       \\
 & ICNet (ECCV2018)~\cite{ferrari_icnet_2018}                     &   -   & 69.5          &   30          & 30.3 @ Titan X  & 49.7               &   -               \\
 & GUNet (BMVC2018)~\cite{mazzini2018guided}                     &    -  & 70.4          &   -          & 33.3 @ Titan Xp  & 29.7               &   -               \\
 & ERFNet (TITS2017)~\cite{romera_erfnet_2018}                       & 72.7 &    69.7       & 27.7         & 11.2 @ Titan X  & 18.4                   &      -            \\ 

\bottomrule

\end{tabular}
\end{table*}

\begin{table*}[tb]
\scriptsize \setlength{\tabcolsep}{2pt}
\caption{IoU per class on the Cityscapes validation set: our method improves the IoU score for most classes compared to lower-resolution baselines with similar complexity. Classes such as person, rider, car, bus and train benefit more from high-resolution processing. The percentage of pixels processed in high-resolution, given per class, shows that our method mostly processes road and sky regions at low resolution. } 
\label{tab:cityscapes_classes}
\centering
\begin{tabular}{lllllllllllllllllllll}
\toprule
{} & {road} & {side-} & {buil-} & {wall} & {fence} & {pole} & {traffic} & {traffic} & {vege-} & {terrain} & {sky} & {per-} & {rider} & {car} & {truck} & {bus} & {train} & {motor-} & {bicycle} & mIoU\\
{} & {} & { walk} & {ding} & {} & {} & {} & {light} & {sign} & {tation} & {} & {} & {son} & {} & {} & {} & {} & {} & {cycle} & {}  & {}
\\ \midrule

{SwiftNet-RN18 }   & 97.8          & 82.9              & 91.8              & 50.9          & 56.2           & 63.0          & 69.0                   & 78.1                  & 92.2                & 61.4             & 94.7         & 80.5            & 60.0           & 94.8         & 78.3           & 84.3         & 75.0           & 6.5                & 76.3  &      76.2    \\ 
{{($2048{\times}1024$)}}       &  &     &   &     &             &        &    &    &    &   &   &   &   &    &   &  &     &   &    &      \\
\midrule

{SwiftNet-RN18}       & \textbf{97.8}   & \textbf{83.0}                 & 91.5                & 54.1            & 52.8             & 60.5            & 65.5                             & \textbf{76.4}                   & 91.9                            & \textbf{62.1}      & 94.6           & 79.2              & 58.2             & 94.2           & 75.5             & 80.5           & 69.6             & 57.6                            & 74.9     &      74.7    \\ 
{($1536{\times}768$)}       &  &     &   &     &             &        &    &    &    &   &   &   &   &    &   &  &     &   &    &      \\
{SegBlocks-RN18 } & \textbf{97.8}   & 82.9                          & \textbf{91.9}       & \textbf{59.5}   & \textbf{57.6}    & \textbf{61.1}   & \textbf{67.0}                    & 76.2                            & \textbf{92.1}                   & 61.9               & \textbf{94.4}  & \textbf{80.4}     & \textbf{59.9}    & \textbf{94.4}  & \textbf{77.2}    & \textbf{82.4}  & \textbf{77.2}    & \textbf{60.1}                   & \textbf{75.7}  & \textbf{76.3}  \\
{($\tau$=0.4)}       &  &     &   &     &             &        &    &    &    &   &   &   &   &    &   &  &     &   &        &  \\

{\quad \textit{\% high-res pixels}}       & \textit{ 8.5} & \textit{45.8} & \textit{64.0} & \textit{67.0} & \textit{88.5} & \textit{86.3} & \textit{85.6} & \textit{82.2} & \textit{59.7} & \textit{68.2} & \textit{36.7} & \textit{93.3} & \textit{96.2} & \textit{66.2} & \textit{61.7} & \textit{68.9} & \textit{85.0} & \textit{90.9} & \textit{93.2}   & -   \\  \bottomrule
\end{tabular}
\end{table*}

\begin{table}[tb]
\scriptsize
\centering
\caption{Memory analysis for SegBlocks-RN18 running in 16-bit floating point, as reported by PyTorch. By storing some image areas at lower resolution, SegBlocks can reduce the memory usage, especially when using larger batch sizes.}
\label{tab:memory}
\begin{tabular}{@{}l|ccc@{}}
\toprule
\textbf{model} & \multicolumn{3}{c}{\textbf{Batch size}} \\
               & 1           & 2           & 4           \\ \midrule
SwiftNet-RN18 (static)         & 480 MB        & 880 MB         & 1681 MB        \\
SegBlocks-RN18 ($\theta = 0.4$)   & 571 MB        & 901 MB         & 1586 MB        \\
SegBlocks-RN18 ($\theta = 0.2$)   & 340 MB        & 721 MB         & 940 MB         \\ \bottomrule
\end{tabular}
\end{table}

\subsubsection{Dataset and setup}

The Cityscapes~\cite{cordts_cityscapes_2016} dataset for semantic segmentation consists of 2975 training, 500 validation and 1525 test images of $2048{\times}1024$ pixels. We use the standard 19 classes for semantic segmentation and do not use the additional coarsely labeled images. 

We train SwiftNet~\cite{orsic_defense_2019} with three different backbones: ResNet-18 (RN18), ResNet-50 (RN50)~\cite{he_deep_2016}, and EfficientNet-Lite1 (EffL1), which corresponds to EfficientNet-B1~\cite{tan2019efficientnet} without squeeze-and-excite and swish modules. Our dynamic processing method is integrated in these baseline networks and we trained models for different $\tau \in [0,1]$, indicating the desired percentage of high-resolution blocks. 
The models are trained on $768{\times}768$ crops, augmented by image scaling between factor 0.5 and 2, slight color jitter and random horizontal flip. The optimizer is Adam, the learning rate is cosine annealed from $4e{{-}4}$  to $1e{{-}6}$ over 350 epochs, weight decay is set to $1e{-4}$ and the batch size is 8. Standard cross entropy loss is used for models with a ResNet backbone, whereas the EfficientNet models use a bootstrapped cross entropy loss~\cite{reed2014training_bootstrap}.  The backbone is pre-trained on ImageNet~\cite{deng_imagenet_2009}. Our method uses a block size of $128{\times}128$ pixels, resulting in a block grid of $16{\times}8$ blocks per image that can adapt the processing resolution to the content. 

\subsubsection{Results and comparison}

Figure~\ref{fig:cityscapes_results} shows the mIoU accuracy of models at various computational costs (GMACs). Static baseline networks of different computational complexity are obtained by adjusting the training and evaluation resolution to $2048{\times}1024$, $1536{\times}768$ and $1344{\times}672$. The complexity of our SegBlocks methods is determined by the number of high-resolution blocks, changed by training for different values of hyperparameter $\tau \in \{0.2,0.4,0.6\}$. 
Our SegBlocks methods consistently outperform the static baseline networks, showing that dynamic resolution processing is more beneficial than sampling all regions at lower resolution. For instance, SegBlocks-RN18 with $\tau{=}0.4$ reduces the computational complexity of the baseline network by $40\%$, without decreasing the mIoU. In contrast, a static baseline network of similar complexity, trained on images of $1536{\times}768$ pixels, achieves 1.7\% lower mIoU.

\begin{figure}[tb!]
\centering
\includegraphics[width=1\linewidth, trim= 0 0 0 0,clip]{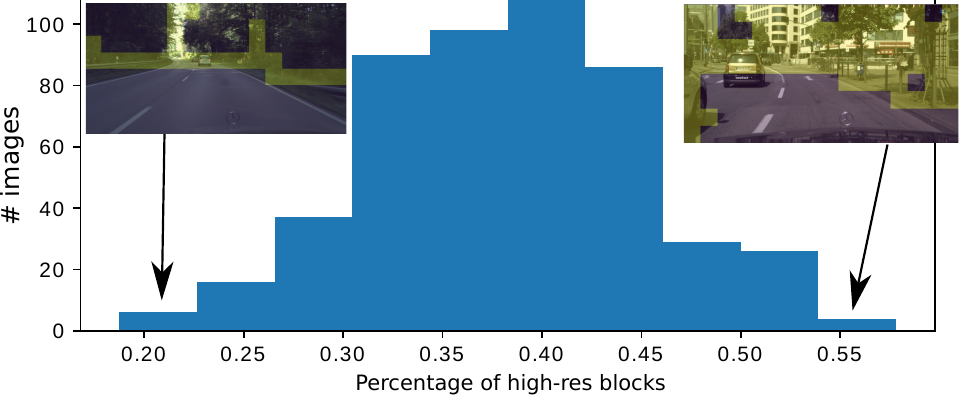}
\caption{Percentage of high-resolution blocks per image, as a histogram over 500 Cityscapes validation images, for SegBlocks-RN18 ($\tau{=}0.4$). Simple images use fewer high-res blocks, down to 20\%, and complex images have up to 55\% high-res blocks (colored yellow).}
\label{fig:cityscapes_distr}
\end{figure}

\begin{figure*}[tb]
\centering
\includegraphics[width=1\linewidth, trim= 0 0 0 0]{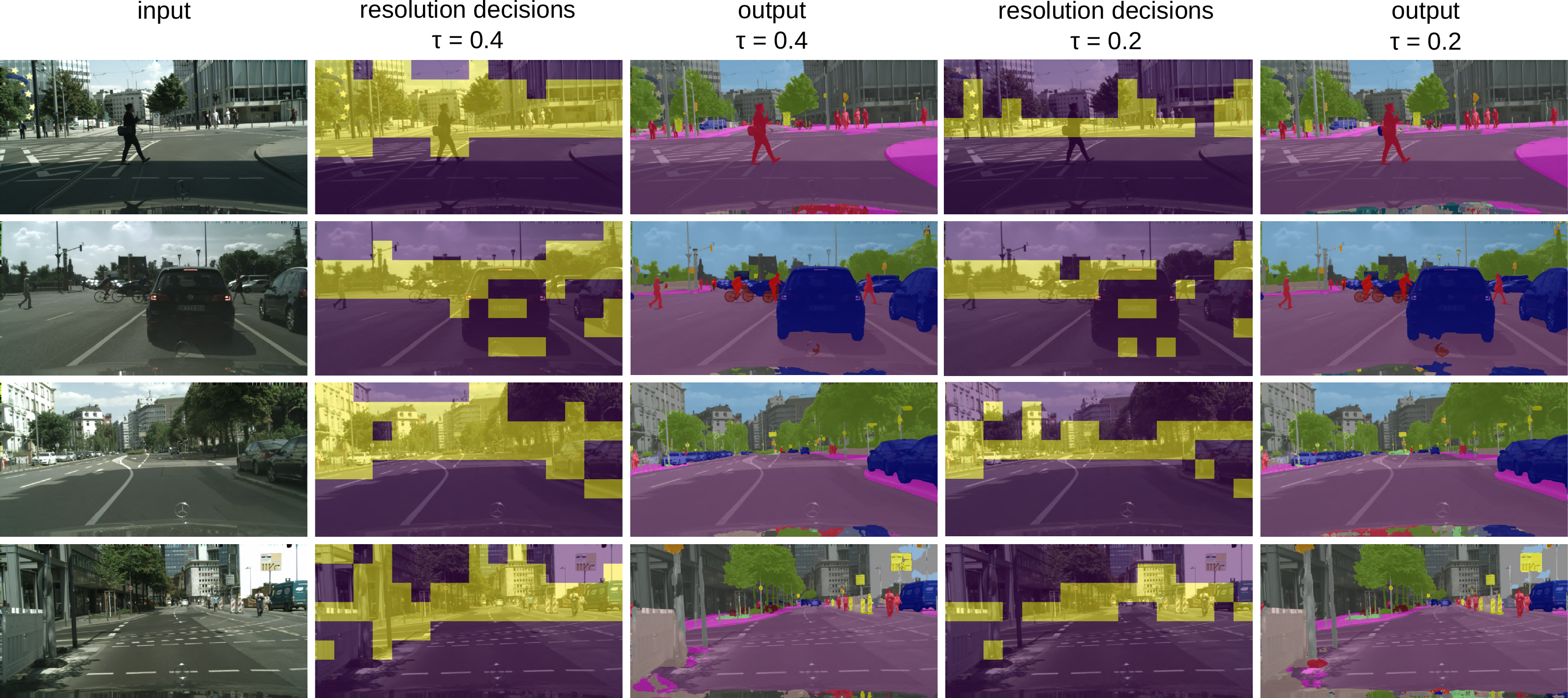}
\caption{Examples of resolution decisions made by the policy network and corresponding segmentation outputs, for SegBlocks-RN18 with $\tau = 0.4$ and $\tau = 0.2$. High-resolution blocks are colored in yellow. 
}
\label{fig:cityscapes_policy_examples}
\end{figure*}

Table~\ref{tab:cityscapes_results} compares the performance of our method with other dynamic methods and non-dynamic efficient segmentation architectures. Compared to other dynamic methods, our method achieves higher mIoU results at faster inference speeds. Patch Proposal Network~\cite{wu_patch_2020} is the most competitive method, and uses an architecture with a global branch and patch-based refinement branch. Our method achieves better accuracy and FPS due to our efficient block-based execution framework. This enables more fine-grained block-based execution, with in total 128 blocks ($16{\times}8$ grid) compared to only 16 large patches, resulting in better adaptability and performance. Similarly, the method of Wu et al.~\cite{wu_real-time_2017} refines regions with a high-resolution branch on large patches. Huang et al.~\cite{huang_uncertainty_2019} use a cascade of two existing network architectures (e.g. lightweight ICNet and accurate PSPNet), where the second network is only applied on complex regions. However, this introduces additional latency and does not re-use any features of the lightweight network, resulting in low efficiency and only 1.8 FPS. The method of Marin et al.~\cite{marin_efficient_2019} (Learning Downsampling) adaptively samples the input image, to more densely sample near segmentation boundaries. Only the theoretical computational cost (GMACs) is reported. HyperSeg~\cite{nirkin2021hyperseg}, based on the EfficientNet backbone~\cite{tan2019efficientnet}, adjusts the weights of the decoder dynamically per image. Stochastic sampling~\cite{xie_spatially_2020} dynamically interpolates features spatially, but does not have an accelerated GPU implementation.

We are also competitive with state of the art architectures for fast semantic segmentation, such as BiSeNet~\cite{ferrari_bisenet_2018} or ERFNet~\cite{romera_erfnet_2018}. It is worthwhile to note that our proposed dynamic resolution method is complementary to further improvements in network architectures.

The table also demonstrates the inference speed improvement achieved using our efficient implementation.
%efficient implementation of the BlockSample, BlockPad and BlockCombine modules.
For instance, our method improves the inference speed of SwiftNet-RN50 from 16 FPS to 30 FPS on a GTX 1080 Ti, achieving real-time performance. This increase of 84 percent is obtained by reducing the theoretical complexity (GMACs) by 63\%,  while the mean IoU is decreased by 1.8\%.
% The inference speed improvement is larger for the ResNet-50 backbones, since its $1{\times}1$ convolutions do not require BlockPad operations.
Memory usage is reported in Table~\ref{tab:memory} as PyTorch's peak allocated memory during inference over the validation set, and indicates that our method also reduces memory consumption by storing low-complexity regions at lower resolution.

Our method adapts the operations to each individual image, and Fig.~\ref{fig:cityscapes_distr} shows the distribution of operations over images.
Qualitative results are shown in Fig.~\ref{fig:cityscapes_policy_examples}, visualizing the resolution decisions of the policy network and the respective segmentation output, for $\tau$ set to 0.4 and 0.2. The policy network, trained with reinforcement learning, properly selects complex regions for high-resolution processing. 

\subsubsection{Per-class analysis}
The IoU result per class is provided in Table~\ref{tab:cityscapes_classes}, as well as the percentage of pixels processed in high-resolution for each class. Our method mainly processes classes such as road, sky and vegetation in low resolution, whereas rider, motorcycle and bicycle are typically sampled in high resolution.

\subsection{CamVid semantic segmentation}

\begin{table}[t]
\scriptsize
\centering
\caption{{Results on CamVid's test set for semantic segmentation~\cite{brostow2009semantic_camvid}. All methods are pretrained on ImageNet~\cite{deng_imagenet_2009}, without Cityscapes pretraining~\cite{cordts_cityscapes_2016}. Results annotated with $\dag$  are determined using open-source implementations. FPS measured on Nivida GTX 1080 Ti.  }}% TODO
\label{tab:results_camvid}
\begin{tabular}{@{}llll@{}}
\toprule
\textbf{Method}              & \textbf{mIoU} & \textbf{GMACs} & \textbf{FPS}                    \\  \midrule
SwiftNet-EffL1~\cite{orsic_defense_2019} (our impl.)    & 75.1          & 13.4           & 48                  \\
SegBlocks-EffL1 ($\tau$ = 0.4)& 74.6  (-0.5\%)& 9.1 (-32\%) & 58  (+20\%) \\
SegBlocks-EffL1 ($\tau$ = 0.2)& 73.4  (-1.7\%)& 6.6 (-50\%) & 64  (+33\%) \\
\midrule

BiSeNet~\cite{ferrari_bisenet_2018}                     & 68.7          &  32.4$^{\dag}$          & 116                  \\
BiSeNetV2~\cite{yu2021bisenet}                    & 72.4          &                & 125                       \\
BiSeNetV2-L~\cite{yu2021bisenet}                  & 73.2          &                & 33                        \\
HyperSeg-S (ResNet-18)~\cite{nirkin2021hyperseg}   & 77.0          &                & 32.5          \\
HyperSeg-S (EfficientNet-B1)~\cite{nirkin2021hyperseg}   & 78.4        &  6.3$^{\dag}$       & 38.0            \\
SFNet (DF2)~\cite{li2020semantic}                  & 70.4          &                & 134                      \\
SFNet (ResNet-18)~\cite{li2020semantic}            & 73.8          &   41.2$^{\dag}$         & 36                      \\                 \bottomrule
\end{tabular}
\end{table}

Another popular benchmark for real-time semantic segmentation is CamVid~\cite{brostow2009semantic_camvid}, having 701 densely annotated frames, divided in 367 training, 101 validation and 233 test images. Following the methodology of related works~\cite{ferrari_bisenet_2018, orsic_defense_2019, yu2021bisenet}, we train on the training and validation subsets and report test scores on 11 semantic classes. We evaluate on images of $960{\times}704$ pixels. We use the same hyperparameters and data augmentations as on Cityscapes, but with crops of 704 pixels. The backbone is pretrained on ImageNet, but we did not pretrain on Cityscapes.
Dynamic methods use a block size of 64 pixels, resulting in a grid of $15{\times}11$ blocks. In order to adjust for the lower input resolution compared to Cityscapes, we remove the stride of the last residual block in the backbone, leading to higher-resolution representations in the spatial pyramid pooling module.

Table~\ref{tab:results_camvid} shows that our results are competitive with other real-time semantic segmentation methods, and our dynamic execution (SegBlocks) improves inference speed over the static baseline network (SwiftNet). However, due to the smaller image dimensions and block size, the speedup is more modest compared to the high-res Cityscapes experiments.

\subsection{Mapillary Vistas semantic segmentation}
Mapillary Vistas~\cite{neuhold_mapillary_2017} is a large dataset containing high-resolution road scene images with labels for semantic and instance segmentation. We use the standard 15000/2000 train/val split, resize all images to $1536{\times}1152$ pixels and use the 65 classes for semantic segmentation. We report validation results, since the test server is not available. SegBlocks usees blocks of 128 pixels, resulting in an adaptive grid of $12{\times}9$ blocks. The hyperparameters are the same as the Cityscapes experiments, with the batchsize reduced to 4 as the number of classes requires more memory, and we train SegBlocks with an EfficientNet-Lite1 backbone for 100 epochs.
Table~\ref{tab:mapillary_results} demonstrates that SegBlocks reduces the number of operations and improves inference speed for the segmentation model on this dataset.

\begin{table}[tb]
\scriptsize
\centering
\caption{Mapillary Vistas validation results. Results with $\dag$ are reported in \cite{hao2021real_wfcdnet} and benchmark FPS with an Nvidia GTX2080 Ti GPU. Other results are benchmarked on an Nvidia GTX1080 Ti.}
\label{tab:mapillary_results}
\begin{tabular}{@{}lrrrr@{}}
\toprule
\textbf{Network}              & \textbf{Input size}                & \textbf{FPS}                & \textbf{GMACs}               & \textbf{mIoU(\%)}            \\ \midrule
SwiftNet-EffL1 (our impl.) & 1536${\times}$1152    &   20.7   &  21.5     &  41.7    \\ 
% SwiftNet-EffL1 (our impl.) & 1024${\times}$768    &   51.1   &  9.6     &  39.8    \\ 
\cdashline{1-5}\noalign{\vskip 0.5ex}
SegBlocks-EffL1 ($\tau$ = 0.4)  & 1536${\times}$1152    &   33.0   &  13.7     &  40.5    \\
SegBlocks-EffL1 ($\tau$ = 0.2)  & 1536${\times}$1152    &   40.1   &  11.5     &  39.8    \\ \midrule
AGLNet~\cite{zhou2020aglnet}    & 2048${\times}$1024 & 53.0   & 24.1G & 30.7 \\
WFDCNet~\cite{hao2021real_wfcdnet}       & 2048${\times}$1024 & 53.1$^\dag$ & 12.5G & 30.5 \\
FPENet~\cite{liu2019feature_fpenet}    & 2048${\times}$1024 & {92.5}$^\dag$ & {3.1G}  & {28.3} \\
DABNet~\cite{li2019dabnet}            & 2048${\times}$1024 & {78.3}$^\dag$ & {20.9G} & {29.6} \\ \bottomrule
\end{tabular}
\end{table}

\subsection{Ablation studies}

\paragraph*{\textbf{Module execution time analysis}}

\begin{table}[tb!]
\scriptsize
\centering
\caption{Time profiling of block modules, as total time taken during the inference of 250 validation images on an Nvidia GTX 1080 Ti GPU.}
\label{tab:timings}
\begin{tabular}{@{}l|cccc@{}}
\toprule
                  & \multicolumn{2}{c}{\textbf{SegBlocks-RN50}} & \multicolumn{2}{c}{\textbf{SegBlocks - RN18}} \\
                      & $\tau$=0.4       & $\tau$=0.2      & $\tau$=0.4        & $\tau$=0.2       \\ \midrule
{mIoU}         & 77.5\%           & 76.2\%          & 76.3\%            & 75.9\%           \\ \cdashline{1-5}\noalign{\vskip 0.5ex}
{GMACs}        & 127.2            & 88.6            & 60.5              & 43.5             \\
{policy GMACs} & 0.34             & 0.34            & 0.34              & 0.34             \\ \cdashline{1-5}\noalign{\vskip 0.5ex}
{Runtime}      & {11.6s}         & {9.3s}         & {5.8s}           & {4.9s}          \\
{Policy Net}   & 0.04s (\textless1\%) & 0.04s (\textless1\%) & 0.04s (\textless1\%) & 0.04s (\textless1\%) \\
{BlockSample}  & 0.12s (1\%)     & 0.13s (1\%)    & 0.11s (2\%)      & 0.12s (2\%)     \\
{BlockPad}     & 0.75s (6\%)     & 0.61s (7\%)    & 0.68s (12\%)     & 0.56s (11\%)    \\
{BlockCombine} & 0.04s (\textless1\%) & 0.03s (\textless1\%) & 0.02s (\textless1\%) & 0.02s (\textless1\%) \\ \bottomrule
\end{tabular}
\end{table}

We profile the time characteristics of our block modules to analyze their overhead. Table~\ref{tab:timings} compares the execution time of the Policy Net, BlockSample, BlockPad and BlockCombine modules with the total runtime. The overhead of the Policy Net, BlockSample and BlockCombine modules is negligible, and the overhead of the BlockPad operation is reasonable with around 10 percent of the total execution time. The BlockPad module has less impact on the ResNet-50 backbone, as its $1{\times}1$ convolutions in the bottleneck function do not require padding. Note that the cost of the BlockPad operation scales with the total number of processed pixels, and thus with the number of high-resolution blocks. 

\begin{figure}[tb]
\centering
\includegraphics[width=0.5\linewidth, trim=2 5 2 1, clip]{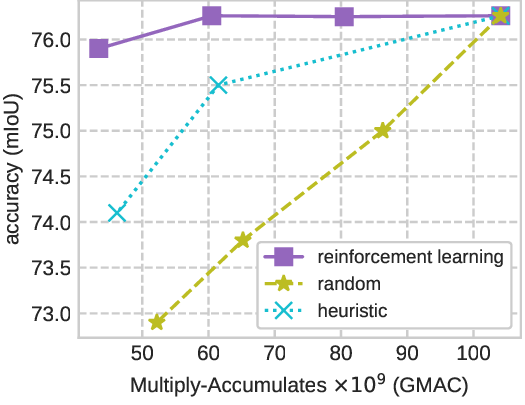}
\caption{
{Comparison of block execution policies. The reinforcement learning policy is proposed in this work. The random policy randomly selects a percentage of blocks per frame for high-resolution processing. The heuristic policy is given in~\cite{verelst2020segblocks}. }
}
\label{fig:policy_ablation}
\end{figure}

\paragraph*{\textbf{Reinforcement learning policy}}
We compare the policy trained using reinforcement learning with other baselines in Fig.~\ref{fig:policy_ablation}. The `random' policy randomly selects blocks for high-resolution processing, with the number of selected blocks given by the target percentage. The heuristic policy was proposed in~\cite{verelst2020segblocks} and selects the regions with the highest visual change, based on the average L2 distance between high- and low-resolution versions of a block. 

\paragraph*{\textbf{Policy network architecture}}
The policy network, determining whether blocks should be executed with high- or low-resolution processing, should be lightweight but powerful enough for the selection task. The ablation (Table~\ref{tab:ablation_policynet}) shows that the size of this network does not significantly impact the results. For our experiments, we use an input resolution of 512x256 with a 4 layer network having 64 features in each layer.

\begin{table}[tb]
\scriptsize
\centering
\caption{Policy network ablation: adjusting the input resolution, number of layers and number of features for the policy network.}
\label{tab:ablation_policynet}
\begin{tabular}{@{}lllll@{}}
\toprule
\textbf{Policy input res.} & \textbf{\# Layers} & \textbf{\# Features} & \textbf{mIoU} & \textbf{Policy GMACs} \\ \midrule
$128{\times}64 $ & 2 & 64  & 75.3 & 0.01 \\
$256{\times}128$ & 3 & 64  & 75.5 & 0.07 \\
$512{\times}256$ & 4 & 64  & 76.3 & 0.34 \\
$1024{\times}512$ & 5 & 64  & 76.1 & 1.51 \\
\midrule

$512{\times}256$ & 4 & 32  & 76.2 & 0.11 \\
$512{\times}256$ & 4 & 64  & 76.3 & 0.34 \\
$512{\times}256$ & 4 & 128 & 76.1 & 1.21 \\ \bottomrule
\end{tabular}
\end{table}

\begin{figure}[tb]
\centering
\includegraphics[width=0.5\linewidth, trim=0 1 20 18, clip]{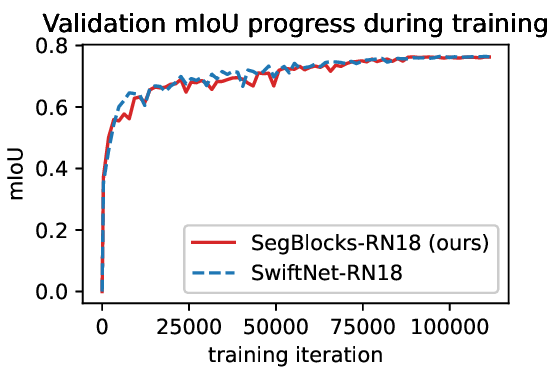}
\caption{Training progress given by validation mIoU for non-dynamic baseline (SwiftNet-RN18) and our dynamic method (SegBlocks-RN18). Learning the policy does not significantly impact the training progress.} %Both methods train at 12 images per second on a GTX 1080 Ti GPU. }

\label{fig:training_progress}
\end{figure}

\paragraph*{\textbf{Training steps and progress}}
We compare the training speed of a standard model with a dynamic SegBlocks model in Fig.~\ref{fig:training_progress}, which shows that the reinforcement learning of the policy does not significantly impact the required number of training steps.

\begin{figure}[!tb]
\centering
  \includegraphics[width=0.32\linewidth]{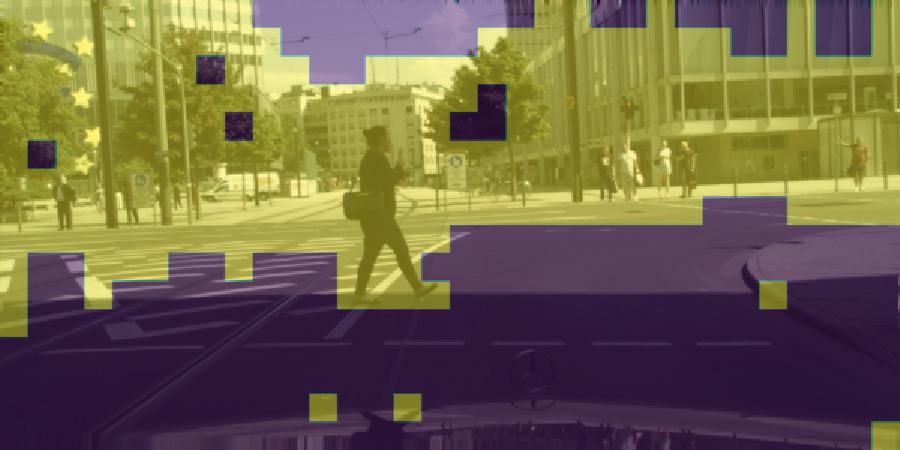}
   \includegraphics[width=0.32\linewidth]{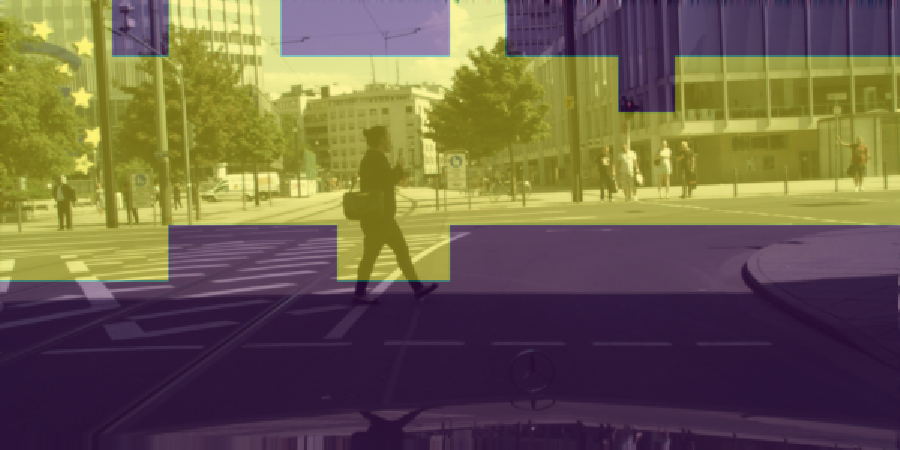}
  \includegraphics[width=0.32\linewidth]{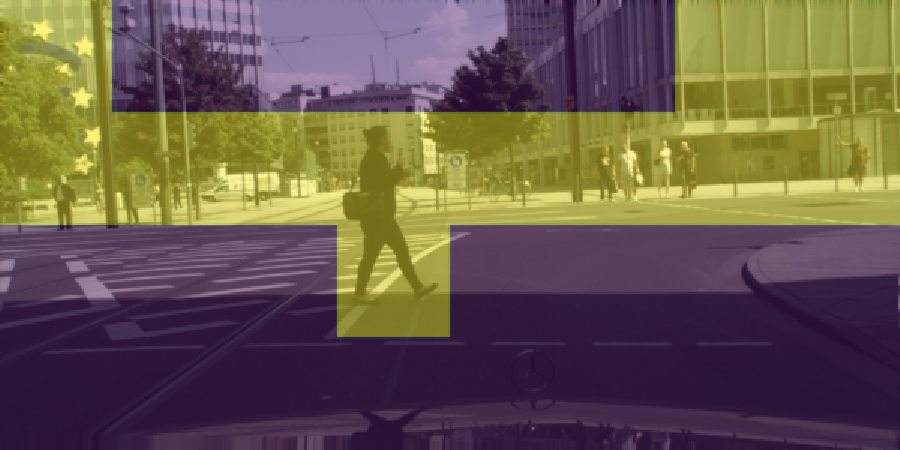}
\caption{Granularity of blocks of 64, 128, and 256 pixels on Cityscapes' images of $2048{\times}1024$ pixels.}
\label{fig:blocksize}
\end{figure}

% \color{black}
\paragraph*{\textbf{Block size}}
Table~\ref{tab:blocksize} compares the impact of the method's block sizes on accuracy and performance. Figure~\ref{fig:blocksize} demonstrates the granularity of each block size on Cityscapes. Larger block sizes offer better inference speeds, by padding fewer pixels and having less overhead, whereas smaller block sizes offer finer control of adaptive processing.

\begin{table}[!tb]
\scriptsize
\caption{Block size comparison for SegBlocks-RN18 with $\tau{=}0.4$ on the Cityscapes validation set \label{tab:blocksize}.}
\centering
\begin{tabular}{@{}lrrrr@{}}
\toprule
\textbf{Block size} & \textbf{mIoU} & \textbf{GMACs} & \textbf{FPS (1050 Ti)} \\ \midrule
$64{\times}64$ & 76.3\% & 59.3 & 11.3 \\
$128{\times}128$ & 76.3\% & 60.5 & 13.5\\
$256{\times}256$ & 75.2\% & 56.4 & 14.6\\
\bottomrule
\end{tabular}
\end{table}

\paragraph*{\textbf{Sampling pattern}}
As described in Section~\ref{sec:blockpad}, the BlockPad module pads individual blocks by sampling their neighbors. When low-resolution blocks are padded with features of high-resolution blocks, these features should be subsampled. Table~\ref{tab:sampling} compares average-sampling with strided subsampling and shows that average-sampling achieves better accuracy (mIoU) with only a slight increase in overhead. In addition, we show that using zero-padding has a significant impact on accuracy, highlighting the importance of BlockPad when processing images in blocks. 

\begin{table}[!tb]
\scriptsize
\caption{Comparison of BlockPad sampling patterns and zero-padding for SegBlocks-RN18 ($\tau{=}0.4$) \label{tab:sampling}}
\centering
\begin{tabular}{@{}lrrrr@{}}
\toprule
\textbf{Method} & \textbf{mIoU} & \textbf{Runtime} & \textbf{BlockPad time} \\ \midrule
Avg-sampling & 76.3\% & 5.8 s & 0.68 s (12\%) \\
Strided sampling & 75.6\% & 5.7 s & 0.63 s (11\%) \\
Zero-padding & 70.2\% & 5.2 s & N.A. (integrated in conv.)\\
\bottomrule
\end{tabular}
\end{table}

\subsubsection{Single residual block analysis}
\begin{figure}[!tb]
\centering
\subfloat[Block size impact]{\includegraphics[width=0.4\linewidth]{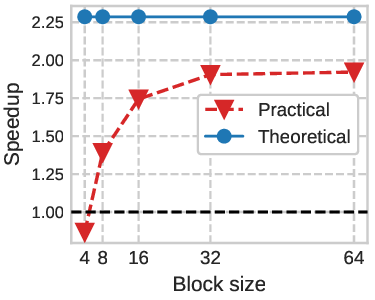}%
\label{fig:speed_vs_blocksize}}
\hfil
\subfloat[Percentage impact]{\includegraphics[width=0.4\linewidth]{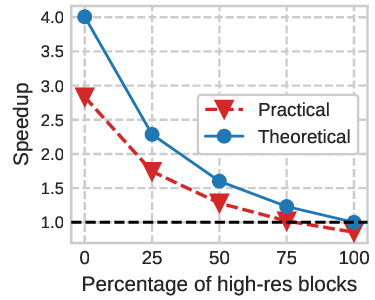}%
\label{fig:speed_vs_percentage}}
\caption{Comparison of theoretical versus practical speedup for a single residual block of ResNet-18, studying the impact of the block size and the percentage of high-resolution blocks on the performance.}
\end{figure}

The overhead introduced by our block modules strongly depends on the block size: larger blocks have fewer pixels in the padding, and require fewer copy operations. Moreover, standard operations, implemented in PyTorch and cuDNN, are often optimized for larger spatial dimensions. We analyze a single residual block to measure the impact of both block size and the percentage of high-resolution blocks.
Figure~\ref{fig:speed_vs_blocksize} studies the impact of the block size, when evaluating half of the blocks at high and half of the blocks at low resolution. The number of floating-point operations is reduced by 56\%, resulting in a theoretical 2.3 times speed increase. In practice, we measure 1.92 times faster inference of the residual block when the block size is larger than 32. Block sizes smaller than 4 pixels result in {slower} inference. Note that, even though block sizes are typically large at the network input (e.g. $128{\times}128$) downsampling in the network reduces the block size by the same factor. The SwiftNet network downsamples by a factor 32 throughout the network, resulting in block sizes of $4{\times}4$ for the deepest network layers.
Figure~\ref{fig:speed_vs_blocksize} shows that, when evaluating using block size 8, the percentage of high-res blocks should be lower than 75\% to achieve practical speedup. For lower percentages, the practical speedup of our implementation is around 70 percent of the theoretical one.

\section{Conclusion}
We proposed a method to reduce the computational cost of existing convolutional neural networks, by evaluating images in blocks and adjusting the processing resolution of each block dynamically. We introduced custom block operations, enabling efficient inference and training of these dynamic neural networks. Our BlockPad module is essential to enable feature propagation between individual blocks. 
A lightweight policy network, trained with reinforcement learning, selects complex regions for high-resolution processing. 
Our method is demonstrated on semantic segmentation of street scenes and we show that the computational complexity of the SwiftNet architecture can be reduced with only a small decrease in accuracy. Dynamically adjusting the processing resolution per block achieves better accuracy than simply downscaling the image to a lower resolution.

The idea of block-based processing and dynamic resolution adaptation can be integrated in various network architectures and computer vision tasks. In particular, our method is suited for dense pixel-wise classification tasks such as depth estimation, instance segmentation or human pose estimation. Furthermore, our block modules such as BlockPad pave the way towards other variants, for instance by processing blocks with varying quantization levels. 
%We also envision specialized implementations of adaptive convolutional neural networks in order to further reduce overhead of these methods. 
Dynamic processing can help to keep the number of operations and energy consumption under control.

\section{Acknowledgments}
This work was funded by FWO on the SBO project with agreement S004418N.

\ifCLASSOPTIONcaptionsoff
  \newpage
\fi



\bibliographystyle{IEEEtran}
\bibliography{references}
\begin{IEEEbiography}[{\includegraphics[width=0.9in,clip,trim= 0 35 0 10]{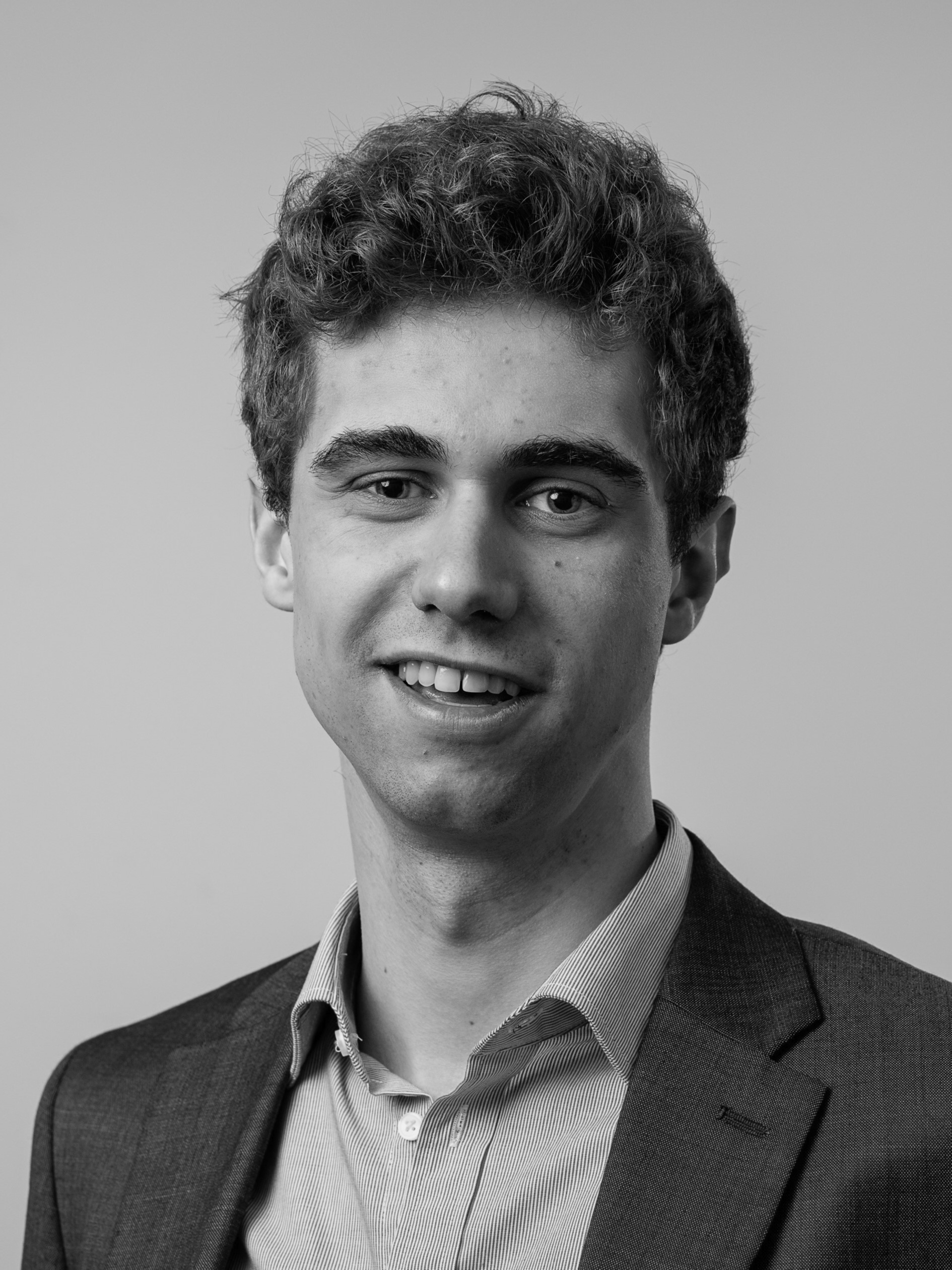}}]{Thomas Verelst} is a Ph.D. student at the Center for Processing Speech and Images (PSI) of KU Leuven university (Belgium). He received a M.Sc. degree in Electrical Engineering at the same university in 2018. His research focuses on efficient and dynamic network architectures for classification, pose estimation and segmentation tasks.
\end{IEEEbiography}

\begin{IEEEbiography}[{\includegraphics[width=0.9in,height=1.05in,clip]{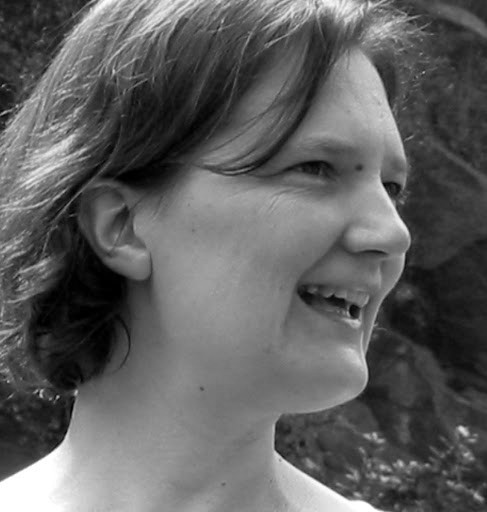}}]{Tinne Tuytelaars} is a full professor at the Electrotechnical department of KU Leuven. Her research focuses on image understanding, with a focus on representation learning, multimodal learning (images and text) and continual learning. In 2009,
she received an ERC starting independent researcher grant, and in 2016, she received the Koenderink award. She has been one of the Program Chairs of the European Conference on Computer Vision 2014 and of IEEE/CVF Conference on Computer Vision and Pattern Recognition  2021, and one of the General Chairs of IEEE/CVF Conference on Computer Vision and Pattern Recognition 2016. She has been associate editor in chief of the IEEE Transactions on Pattern Analysis and Machine Intelligence and serves as area editor for the International Journal on Computer Vision.
\end{IEEEbiography}

\end{document}